\documentclass[conference]{IEEEtran}
\ifCLASSINFOpdf
 
\else
 
\fi

\usepackage{times}
\usepackage{helvet}
\usepackage{courier}
\usepackage[english]{babel}
\usepackage[utf8x]{inputenc}

\usepackage{amsmath}
\usepackage{graphicx}
\usepackage[colorinlistoftodos]{todonotes}
\usepackage[colorlinks=true, allcolors=blue]{hyperref}
\usepackage{algorithmic}
\begin{document}

\title{Deep Transfer Learning for Static Malware Classification}
\author{\IEEEauthorblockN{Li Chen}
\IEEEauthorblockA{Security and Privacy Research, Intel Labs\\
Hillsboro, OR 97124\\
%Email: li.chen@intel.com
}}
\maketitle
\thispagestyle{plain}
\pagestyle{plain}

\begin{abstract}
We propose to apply deep transfer learning from computer vision to static malware classification. In the transfer learning scheme, we borrow knowledge from natural images or objects and apply to the target domain of static malware detection. As a result, training time of deep neural networks is accelerated while high classification performance is still maintained. We demonstrate the effectiveness of our approach on three experiments and show that our proposed method outperforms other classical machine learning methods measured in accuracy, false positive rate, true positive rate and $F_1$ score (in binary classification). We instrument an interpretation component to the algorithm and provide interpretable explanations to enhance security practitioners' trust to the model. 
We further discuss a convex combination scheme of transfer learning and training from scratch for enhanced malware detection, and provide insights of the algorithmic interpretation of vision-based malware classification techniques.
\end{abstract}

\IEEEpeerreviewmaketitle
\section{Introduction}

Malware is a type of software that possesses malicious characteristics to cause damage to the user, computer or network. Categories of malware include virus, trojan horses, worms, spyware, ransomware and so on. Static analysis is a quick and straightforward way to detect malware without executing the application or monitoring the run time behavior. One main technique is the so-called signature matching, where the goal is to search whether the strings in the code actually match any identified malicious patterns in database. However when the code is obfuscated or morphed, signature matching cannot be applied and becomes less resilient to detect malicious patterns. Moreover the rapid increase of signatures, often in exponential growth, makes the signature matching less efficient and scalable. %Additionally, detection schemes that only use malware signatures are less resilient and less effective to identify unknown or morphed malware. 

Machine learning, as a probabilistic way, has demonstrated to be quite powerful for large-scale malware detection. One key ingredient for a successful machine learning malware detector is feature engineering.
%In malware detection, static analysis uses malware binary or disassembly code to understand the malicious properties or activities without executing the application. However 
 For static analysis, often a disassembly step is done on the application binary, so the code becomes human-readable. What follows is heavy emphasis on excellent feature construction including binning the type of function calls, counting the number of loops, and applying language models on the PE headers. 
 %can be natural language processing step, histogram or loop calculation and so on. '
 Such feature processing can result in very high dimensional feature space. For such high-dimensional dataset, dimension reduction or feature selection are used, but the number of dimensions to select or the procedure of feature selection is generally a challenging problem in both practical and theoretical machine learning \cite{blum1997selection,langley1994selection}. Hence an automated and effective method without much manual efforts of feature construction is greatly desired. 

In this paper, we propose an enhanced malware detection framework employing deep transfer learning to train directly on malware images. Our approach is motivated by visual inspection of application binaries plotted as grey-scale images: there exists textural and structural similarities among malware from the same family and dissimilarities between malware and benignware as well as across different malware families. This observation is also seen in \cite{makandar2015malware, niemela2014malware, nataraj2011malware}.

Naturally we consider malware classification as a computer vision problem. Our proposed method is able to consume the entire application binary with or without code obfuscation and independent of signatures. Further, it does not require manual feature engineering or extraction effort. 

The step of transfer learning step is employed to borrow source knowledge from large scale of natural images or objects and applied on the target task of malware detection. Compared to existing vision-based malware classification approaches \cite{nataraj2011malware, makandar2016texture, makandar2015malware,su2018lightweight}, the transfer learning scheme greatly accelerates the training of very deep neural networks while maintaining high classification performance even on smaller sized dataset.

In our experiments, we show that our method obtains the highest classification accuracy, lowest false positive rate, highest true positive rate and highest $F_1$ score compared to classical machine learning algorithms such as shallow neural networks, naive Bayes \cite{lewis1998naive}, $k$-nearest neighbor \cite{cover1967nearest}, linear discriminant analysis\cite{fisher1936use}, random forest\cite{breiman2001random}, XGB\cite{chen2016xgboost}, and support vector machine \cite{steinwart2008support} with linear and radial kernels respectively. Furthermore, we compared transfer learning with training from scratch and demonstrate the advantage of using transfer learning on limited amount of dataset to achieve superior performance.  
 
Despite the classification effectiveness of our proposed method, understanding the reason why the deep transfer learning model makes such predictions on the malware images is critical for security researchers and practitioners. The interpretations will generate valuable insights to triage malware families and enhance the practitioners' trust to the model. Hence an effective model for deployment not only has the best classification performance but also provides explainable interpretation for its predictions. Our proposed method addresses the model interpretability and trustworthiness aspect via utilizing the local-interpretable model-agnostic explanation approach \cite{ribeiro2016should}. We illustrate in Section \ref{sec:exp} using the image-based deep learning classifier for intelligent interpretability. Such interpretability again highlights the advantage of deep transfer learning classification on malware image classification, so that the algorithm is no longer black-box.

The contributions of our paper are summarized as below:
\begin{itemize}
    \item We propose using transfer learning from computer vision to apply for static malware classification. The transfer learning scheme greatly accelerates training deep neural networks while maintaining high classification performance.
    \item We thoroughly examine and evaluate the effectiveness of the proposed method in real-world experiments. Our method achieves the best accuracy at 99.25\%, 98.13\% and 99.67\% respectively in three datasets, compared with training from scratch scheme and classical but effective machine learning algorithms. When evaluated in false positive rate, true positive rate and $F_1$ score, our method also outperforms other algorithms in consideration. 
    \item We provide interpretable explanations of the deep transfer learning for malware image classification. The interpretation component of our approach adds trust to the model and becomes essential for security practitioners to put faith for model deployment. 
\end{itemize}

The rest of the paper is organized as follows. Section \ref{sec:related_work} provides background on image-based malware classification approaches. Section \ref{sec:method} presents our proposed methodology in detail. Section \ref{sec:exp} applies our method on real datasets and demonstrates its superior  performance compared with classical but effective machine learning classifiers as well as training from scratch. We also illustrate how to generate reasonable and interpretable explanations of our approach to better understand the model prediction and enhance trust in the model. Section \ref{sec:disc} summarizes the paper and discusses future research directions. 
\section{Related Work} \label{sec:related_work}

For static malware classification, signature matching is effective at identifying known malware, but new or morphed malware code not present in the signature database can easily evade detection. Machine learning based malware detection provides an automatic and probabilistic way to detect malware. Particularly these models extend the effectiveness to detect malware with great scalability and efficiency. 

Classification on image-based malware data has shown to be very promising. Authors in \cite{nataraj2011malware} were the first to propose converting malware binaries into digital grey-scale images. They applied GIST feature extraction on the malware images followed by k-nearest neighbor with Euclidean distance and achieved 98\% classification accuracy on a malware benchmark consisting of 25  malware families. We use the same dataset for our experiment in Section \ref{sec:exp} and demonstrate our proposed method has higher classification accuracy. 
In \cite{makandar2015malware}, the authors proposed using artificial neural network to perform malware classification. They apply wavelet transformation for feature extraction and then train support vector machine and artificial neural network on the extracted features. In \cite{yue2017imbalanced}, the author proposed to use convolutional neural network for multi-family malware classification. In \cite{su2018lightweight}, the authors proposed to use malware images for IoT malware classification. In \cite{chen2018henet}, the authors proposed a hierarchical ensemble neural network on Intel\textsuperscript{\tiny\textregistered} Processor Trace for effective exploit detection. The collected trace is converted into time series of images and deep convolutional neural network is trained on the images representing the dynamic behaviors of the malware.

\section{Methodology} \label{sec:method}
The motivation of our approach comes from visual inspection of the malware binaries represented as grey-scale images and observe that the malware from the same family share structural similarities and malware from different families show distinct structural or textural information. 
%Visual inspection on the grey-scale malware images demonstrates the malwares from the same family share structural similarities and malwares from different families show distinct structrual or textural information.
%similarities among malware or attacks of the same family and dissimilarities among malwares from the different families \cite{chen2018henet, nataraj2011malware, niemela2014malware, makandar2015malware, shaid2014malware}. 
Such visual inspection provides foundation for treating the malware classification problem as a vision classification task. 
%our approach to borrow domain knowledge learned from natural images and fine-tune on the target domain of malware classification. %Authors in \cite{niemela2014malware, makandar2015malware, shaid2014malware} have similar observations and thus apply computer vision techniques to perform malware classification.

Deep learning has demonstrated state-of-the-art performance on large-scale image classification. Particularly, transfer learning has heavily been employed in computer vision. The idea of transfer learning is to borrow the knowledge learned from a model used in one domain and apply it to another targeted domain. Typically practitioners would take a pre-trained model from a type of image dataset, freeze a portion of the layers and fine-tune the last few layers on the newly obtained dataset. The advantages of transfer learning include accelerating training time, reducing parameters and architecture search for deep neural networks and maintaining high classification performance especially on relatively smaller-sized dataset. 

\begin{figure*}
\centering
\includegraphics[width = 0.8\textwidth]{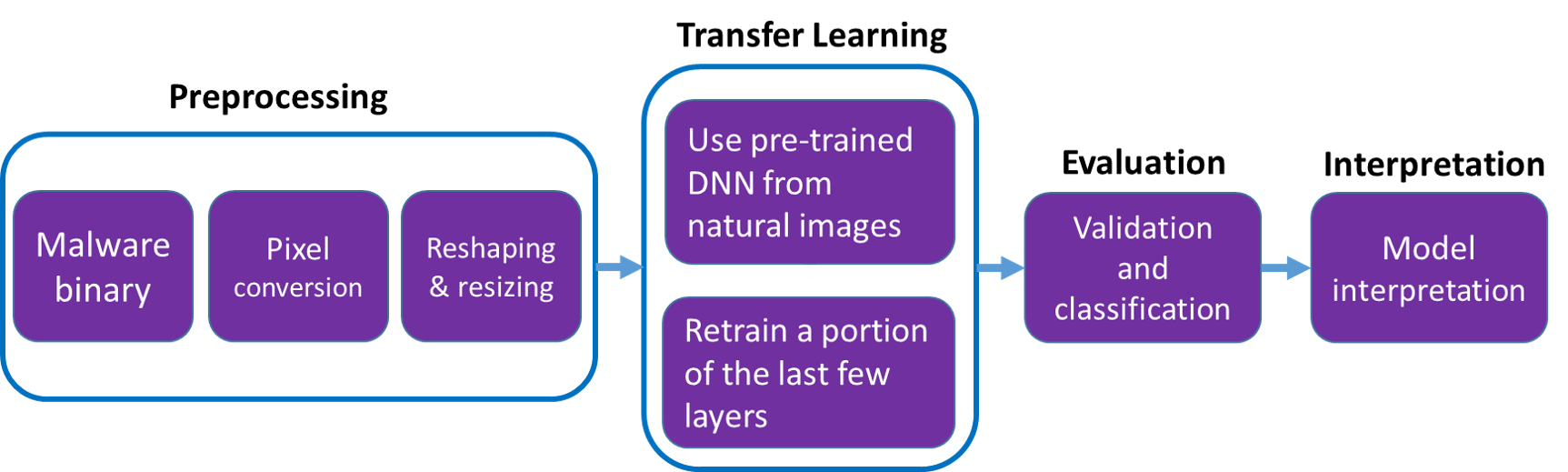}
\caption{\label{fig:method}An overview of our proposed method, which utilizes transfer learning on malware images and provides visually explainable interpretation. }
\end{figure*}

Our proposed method leverages the value of transfer learning to train a highly effective malware classifier for static malware classification. Essentially our methods consists of four main steps: reshaping, resizing and replicating as preprocessing, training deep neural network via transfer learning, evaluation and interpretation. The preprocessing step directly converts the raw binary into two-dimensional images with minimum feature engineering. Subsequently transfer learning in computer vision are applied on the malware images. The interpretation component offers explainability on predictions and thus helps the security practitioners trust the model for deployment. We describe our steps in detail below. Figure \ref{fig:method} is an overview of our proposed framework.

%\begin{figure}
%\centering
%\includegraphics[width=0.5\textwidth]{Figs/system_overview.png}
%\caption{\label{fig:method}The overview of utilizing vision-based transfer learning for malware %analysis.}
%\end{figure}

%There are two learning components in our proposed framework: transfer learning and an optimal convex combination network. In the transfer learning component, the deep features are automatically extracted and learned from natural objects and images and then fine-tuned on the target task of malware analysis. In the convex combination neural network scheme, the linear combination of transfer learning and training from scratch demonstrate to provide either higher accuracy or lower false positive rate than either training scheme alone. 

\subsection{Preprocessing}
Given a static binary, we map it directly to an array of integers between 0 and 255. Hence each binary is converted into a one-dimensional array $v \in [0, 255]$. Then the array $v$ is normalized to $[0,1]$ by dividing by 255.

The normalized array $v$ is then reshaped into a  two-dimensional array $v'$, as seen in Table \ref{tab:reshaping}. We follow the resizing as in \cite{nataraj2011malware} where the width is determined with respect to the file size.
%our reshaping table is a function of the distribution of the file sizes. 
The height of the file is the total length of the one-dimensional array divided by the width. We round up the height and pad zeros if the width is not divisible by the file size.

%We convert a given binary into a vector of 8-bit unsigned integers, and then reshape it into a two-dimensional array. We modify the reshaping table described in (cite) and our modified reshaping table is seen in Table \ref{tab:reshaping}. 

%When the distribution of file size is very spread out, we remove files with size less than 50 kb, since after reshaping, the patterns are greatly distorted. For such cases, we recommend a reshaping table with even finer interval segmentation of file sizes. Figure \ref{fig:bad_file_example} shows an example of binary with file size less than 50 kb. 

We resize the grey scale images into $m$ by $m$ matrix. The choice of $m$ is typically 224 or 229 to fit the input size of the pre-trained models via ImageNet.
%depends on whether one uses transfer learning or CC-Net, which we will explain in Section and . 
After resizing the two-dimensional arrays into squared gray-scale images, we replicate the channel three times to convert it into RGB images.

\begin{table}
\begin{center}
\begin{tabular}{ |c|c|c| } 
 \hline
File size (kb) & Width  \\ \hline 
 $(0, 10]$ & 32 or remove samples \\ 
 $(10, 30]$ & 64\\ 
 $(30, 60]$ & 128 \\ 
 $(60, 100]$ & 256  \\ 
  $(100, 200]$ & 384  \\ 
  $(200, 500]$ & 512  \\ 
  $(500, 1000]$ & 768 \\ 
  $(1000, 2000]$ & 1024  \\ 

 $>200$ & 2048  \\ 

 \hline
\end{tabular}
\end{center}
\caption{We reshape the one-dimensional array $v$ into two-dimensional array $v'$ using the above relationship between file size and width. }
\label{tab:reshaping}
\end{table}

%We modify a reshaping table as in (cite) to transform convert the one-dimensional array into two-dimensional array. 

%Transfer learning has heavily been used in the computer vision domain. 
%Transfer learning in computer vision 

%Task to task

%Our visual inspection on malware datasets provides evidence that malware from different families have different patterns. 

%Our proposed framework has the following procedure. 

%Then we follow a modification of a rule-of-thumb reshaping table in (cite), and convert the one-dimensional array into two-dimensional array. We propose to use either 1) transfer learning, which learned from natural image objects and applies the knowledge to the target domain malware analysis, or 2)a convex combination neural network to optimize the combination of training from scratch and transfer learning for enhanced classification performance. We will describe how both methods proceed for malware classification in Section and . 

%We directly map the malware binaries as a 1-dimensional array consisting of integers between 0 and 255, then reshaping it to convert into 2-dimensional images. 

%\subsection{Reshaping and Resizing}

\subsection{Training}
The second step is to conduct transfer learning on the reshaped and resized images, where a portion of the layers are frozen and the last few layers are retrained on the malware images. Because of the transfer learning scheme, our framework is flexible to support a variety of deep learning neural networks such as Inception \cite{szegedy2016rethinking, szegedy2017inception}, VGG \cite{simonyan2014very}, ResNet \cite{he2016deep}, DenseNet \cite{huang2017densely}. Using transfer learning, we save significant amount of time searching for the neural network architectures, parameters and optimizers. 
%Figure \ref{fig:transfer_learning} is an example of fine-tuning procedure on Inception-v1 neural network. 

Given the same data representation as RGB images and the same amount of least feature engineering efforts, our proposed approach outperforms popular malware classifiers such as shallow neural networks, naive Bayes \cite{lewis1998naive}, $k$-nearest neighbor \cite{cover1967nearest}, linear discriminant analysis\cite{fisher1936use}, random forest\cite{breiman2001random}, XGB\cite{chen2016xgboost}, and support vector machine \cite{steinwart2008support} with linear and radial kernels respectively. Additionally, we compared transfer learning with training from scratch. Since we have much fewer malware images as opposed the size of ImageNet \cite{deng2009imagenet}, training from scratch on the large deep neural networks may not converge. Thus for training from scratch, we use the small versions of the deep neural networks for comparison. On the other hand, this emphasizes the advantage of using transfer learning when data size or amount is limited.

\subsection{Evaluation}
We use classification accuracy, false positive rate, true positive rate, $F_1$ score (in binary classification) as evaluation metrics for our proposed methodology. We will show in Section \ref{sec:exp} that our proposed method outperforms all other selected machine learning algorithms in terms of all these metrics on real-world datasets. 

\subsection{Interpretation}
While classification performance is one key metric for model selection in malware classification, solely relying on classification power without explainablity or interpretation puts blind trust on the model, and that model may not even perform well in the wild. Especially for security researchers and practitioners, %an effective malware classifier in deployment means, 
in addition to highly accurate predictions, they must be able to understand why the model makes such predictions and interpret the explanations as faithful and intelligent in order to trust the model.

%the explanations for the predictions would make sense as to extend trust to the model. 

%blindly putting faith in the best performing model alarms the security practitioners
To provide interpretation, we essentially follow the local-interpretable model-agnostic explanation algorithm in \cite{ribeiro2016should}. We first represent the malware images via super pixels, which  are regions or patches of pixels adjacent to each other. The super-pixel representation becomes an visually interpretable feature on the malware images. Then a binary vector of \{0, 1\}-values is imposed on each super pixel, where 0 indicates the absence of the super pixel and 1 indicates the presence of the super pixel. We use sparse linear classifiers to train on the binary vectors and obtain the trained coefficients. The positive coefficients associated with each super pixel means the regions of pixels are contributing to the classification decision by the model. The negative coefficients associated with the super pixels mean these regions in the malware image do not contribute to the classification decision. We will illustrate in Section \ref{sec:exp} how to use this interpretation component to understand the prediction reason. Security practitioners can identify the locations of interest generated by the interpretation and refer back to the binary file to identify new patterns or signatures. 

%Perfect combination of security domain expertise and machine learning. 

%Our image-based representation enables visual interpretability for deep transfer learning malware classifier. interpretability and understanding of the models. We use a model-agnostic approach established in \cite{ribeiro2016should} to provide interpretability of our proposed methodology and evaluate the trustworthiness of the model. %In our experiment session, we show that our proposed approach achieves the best trustworthiness. 
\section{Experiments} \label{sec:exp}
In this section, we apply our proposed method on three real datasets and demonstrate our proposed method achieves superior classification performance. For all the experiments, we use the pre-trained Inception-V1 on ImageNet at 126-th epoch for transfer learning, where we freeze all the layers before the last pooling layer and retrain the fully connected layer using the malware datasets. We compare Inception transfer learning with training from scratch using the small Inception neural network. The classical machine learning algorithms used for comparison are shallow fully connected neural networks (shallow NN), naive Bayes, 5-nearest neighbor (5NN), linear discriminant analysis (LDA), random forest, XGB, support vector machine with linear kernel (SVM-linear), support vector machine with radial kernel (SVM-radial), all of which are applied on the pixel representation of the malware images for fair comparison. We consider both array-based and vectorization-based representation of the malware image datasets.  

Experiment I and II are multi-family malware classification, where we will report the classification accuracy, average false positive rate and average true positive rate as the metrics for comparing the effectiveness of the classifiers. The average false positive rate is defined as 

\begin{equation}
    \overline{FPR} := \frac{1}{n}\sum_{i=1}^nfpr_i, 
\end{equation}
where $n$ is the total number of malware families and $fpr_i$ denotes the false positive rate for the sample from any other malware family being misclassified as from the $i$-th malware family. Similarly, the average true positive rate is defined as 
\begin{equation}
    \overline{TPR} := \frac{1}{n}\sum_{i=1}^ntpr_i,
\end{equation} 
where $n$ is the total number of malware families and $tpr_i$ denotes the true positive rate for the sample from $i$-th malware family being correctly classified as from the $i$-th malware family. 
Experiment III is a binary-class classification problem on benign and malware samples, where we will report the accuracy, false positive rate, true positive rate and $F_1$-score on test set for algorithm comparison.

%We report the accuracy, flase positive rate, true positive rate and area under the curve for classificatioon performance. 
\subsection{Multi-family malware classification I}

We evaluate our method on a benchmark malware dataset used in \cite{nataraj2011malware, niemela2014malware, makandar2016texture, yue2017imbalanced}. There are 9458 malware samples from 25 malware families: Allaple.L, Allaple.A, Yuner.A, Lolyda.AA 1, Lolyda.AA 2, Lolyda.AA 3, C2Lop.P, C2Lop.gen!g, Instantaccess, Swizzot.gen!I, Swizzot.gen!E, VB.AT, Fakerean, Alueron.gen!J, Malex.gen!J, Lolyda.AT, Adialer.C, Wintrim.BX, Dialplatform.B, Dontovo.A, Obfuscator.AD, Agent.FYI, Autorun.K, Rbot!gen, Skintrim.N. We randomly split the training:valiation:testing ratio to be 0.8:0.1:0.1. We first resize the malware images into $224\times224$ grey-scale images and then replicate the channel three times so the images are RGB of size $224 \times 224 \times 3$.

%The reported accuracy from paper cite is , from paper cite is. 

\begin{figure}
\centering
\includegraphics[width=0.45\textwidth]{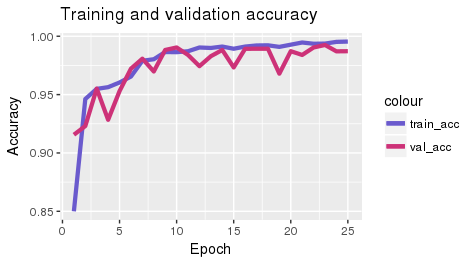}
\caption{\label{fig:ucsb_training_val_acc}The training and validation accuracy of our method over 25 epochs for multi-family malware classification. To avoid overfitting, we use the model at the 14-th epoch for prediction on test set.}
\end{figure}

\begin{figure}
\centering
\includegraphics[width=0.45\textwidth]{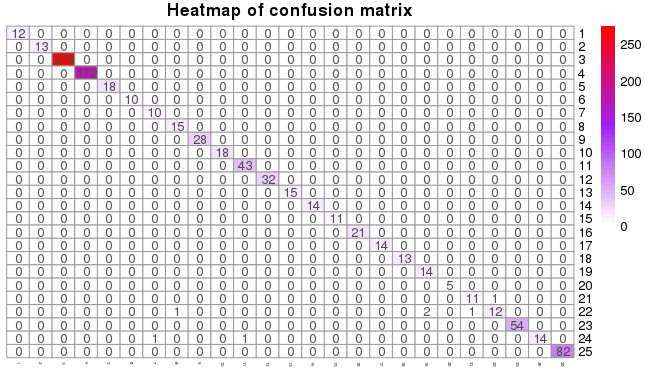}
\caption{\label{fig:confusion_ucsb}The confusion matrix of our proposed method on the test set. The classification accuracy by our proposed method is 99.25\% with average false positive rate at $0.030\%$ and average true positive rate at $98.15\%$ respectively.}
\end{figure}

We train the last fully connected layer of the network for 25 epochs and plot the training and validation accuracy in Figure \ref{fig:ucsb_training_val_acc}. Due to the bias-variance trade-off\cite{duda2012pattern}, we select the model trained at the 14-th epoch and apply it on the test data. On the test set, the classification accuracy by our proposed method is 99.25\% with average false positive rate at $0.030\%$ and average true positive rate at $98.15\%$ respectively. These three metrics outperform all the other 17 classifiers as shown in Table \ref{tab:dataset1}. The confusion matrix is plotted in Figure \ref{fig:confusion_ucsb}.

When applying the selected classical machine learning algorithms on images of 224 $\times$ 224, the dimension is 50176, which is much greater than 7481 the number of training samples. Due to the curse of dimensionality \cite{duda2012pattern}, we first apply principal component analysis (PCA) on the training set and select the first 100 Principal Component (PC) dimensions based on the scree plot. Then PCA rotation is applied on the test matrix to obtain the low-rank representation. XGB performs the best second after our method with accuracy of 98.44\%, average FPR at 0.065\% and average TPR at 95.97\%, while naive Bayes performs the worst with accuracy of 88.05\%, average FPR at 0.501\% and average TPR at 88.84\%.

Training from scratch is done using small Inception neural network where the images are resized to 28 by 28 with single channel, since we do not have enough large dataset to enable training from scratch for large networks. The resulted accuracy, average FPR and average TPR are  95.28\%, 0.211\% and 87.68\% respectively, which are less effective compared to our method. This could be due to information loss from resizing. We also apply the selected classical algorithms on the images vectorized into one-dimensional arrays of length 784. Random forest and XGB have competitive performance as training from scratch. Surprisingly, the performance of 5NN degrades significantly. Resizing and $L_2$ distance could be contributing to the degradation. 

We note the performance of the other classifiers in comparison might improve if we construct features rather than image representation. However this again highlights the advantage of deep learning of saving cost at feature engineering.

\begin{table*}[t]
\begin{center}
\begin{tabular}{ |l|c|c| c| l| } 
 \hline
Algorithm & Accuracy & $\overline{FPR}$ & $\overline{TPR}$ & Data shape\\ \hline 
Our method & \textbf{99.25\%} & \textbf{0.030\% } & \textbf{98.15\%} &  $224 \times 224 \times 3$ \\ 

%TFS Small Inception $\circ$ PCA & 95.28\% & 0.211\% & 87.68\% &   $224 \times 224 \times 1 \rightarrow 50176 \rightarrow 100$\\ 
TFS via shallow NN $\circ$ PCA & 97.14\% & 0.120\% & 91.78\% & $224 \times 224 \times 1 \rightarrow 50176 \rightarrow 100$\\ 

Naive Bayes$\circ$ PCA  & 88.05\%& 0.501\% &88.84\%&$224 \times 224 \times 1 \rightarrow 50176 \rightarrow 100$\\ 
5-nearest neighbor$\circ$ PCA  & 97.90\%& 0.087\% & 94.79\%&$224 \times 224 \times 1 \rightarrow 50176 \rightarrow 100$\\ 
LDA $\circ$ PCA & 92.51\% & 0.334\%  & 83.18\% &$224 \times 224 \times 1 \rightarrow 50176 \rightarrow 100$ \\ 
Random forest$\circ$ PCA  & 98.12\%& 0.078\% &  95.14\%&$224 \times 224 \times 1 \rightarrow 50176 \rightarrow 100$ \\ 
XGB $\circ$ PCA & 98.44\%& 0.065\%  &  95.97\%&$224 \times 224 \times 1 \rightarrow 50176 \rightarrow 100$\\ 

SVM-linear$\circ$ PCA  & 97.74\% & 0.095\%  & 94.30\% &$224 \times 224 \times 1 \rightarrow 50176 \rightarrow 100$\\ 
SVM-radial$\circ$ PCA  & 95.69\% & 0.179\% &  90.02\%&$224 \times 224 \times 1 \rightarrow 50176 \rightarrow 100$ \\ 
TFS via Small Inception & 95.28\% & 0.211\% & 87.68\% &   $28 \times 28 \times 1$\\ 
TFS via shallow NN & 93.00\% & 0.303\% &81.91\% &$28 \times 28 \times 1 \rightarrow 784$\\ 

Naive Bayes & 94.02\%& 0.249\% &85.65\%&$28 \times 28 \times 1 \rightarrow 784$\\ 
5-nearest neighbor & 44.40\%& 2.257\% & 56.37\%&$28 \times 28 \times 1 \rightarrow 784$\\ 
LDA $\circ$ PCA & 91.07\% &0.384\%  & 78.71\% &$28 \times 28 \times 1 \rightarrow 784 \rightarrow 50$\\ 
Random forest & 95.53\%& 0.199\% &  85.71\% & $28 \times 28 \times 1 \rightarrow 784$ \\ 
XGB & 95.37\%& 0.192\%  &  86.37\%&$28 \times 28 \times 1 \rightarrow 784$\\ 

SVM-linear & 92.14\% & 0.379\%  & 78.12\% &$28 \times 28 \times 1 \rightarrow 784$\\ 
SVM-radial & 92.25\% & 0.374\% &  78.36\% & $28 \times 28 \times 1 \rightarrow 784$ \\ 
\hline
\end{tabular}
\end{center}
\caption{Comparison of algorithm performance on Dataset I. Our proposed method achieves the highest classification accuracy, highest average true positive rate and lowest false positive rate. }
\label{tab:dataset1}
\end{table*}

\subsection{Multi-family malware classification II}
The Microsoft Malware Dataset\cite{ronen2018microsoft} was a dataset used in the Microsoft Malware Classification Challenge in 2015. This dataset is almost half a terabyte when uncompressed and has become a standard benchmark for malware classification research. There are 10868 malware samples from nine malware families: Ramnit, Lollipop, Kelihos\_ver3, Vundo, Simda, Tracur, Kelihos\_ver1, Obfuscator.ACY and Gatak. 

The raw data contains each file's binary content without the header. Following Section \ref{sec:method}, we first reshape the binary into images, then resize the two-dimensional arrays into size of 224 by 224, and duplicate the grey scale three times to convert into three channels finally. Our training : validation : test ratio is set at 0.8:0.1:0.1. 

%We apply transfer learning by first employing the weights of a pre-trained neural network. In this case, we use the pre-train Inception network, which has proven to achive high-classification result on the ImageNet competition. Then we train the last two fully connected layers on our small train set containing the malware binaries. Because the pre-trained neural network is 3-channels while our training data is one-channel, we duplicate the grey scale by three to convert into three channels. 
%Our classification result on the 9-family classification problem is shown as follows. Since this is a multi-class classification problem, we report an accuracy of $98.8\%$, average false positive rate defined same as Equation \ref{eq:avg_fpr} of $0.165\%$ and average true positive rate defined as Equation \ref{eq:avg_tpr}. 
%We further compare with classical machine learning methods. We use the model trained at the 13-th epoch due to bias-variance tradeoff. 

%Training from scratch using small inception, since we do not have enough large dataset to enable training from scratch for large NN network. 

\begin{figure}
\centering
\includegraphics[width=0.45\textwidth]{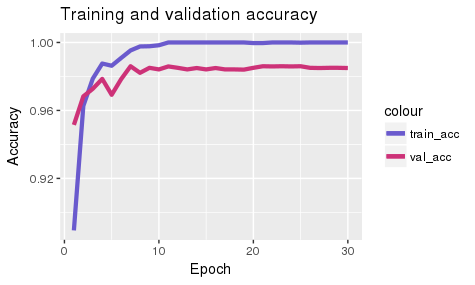}

\caption{\label{fig:training_validation_acc_kaggle}The training and validation accuracy of our method over 30 epochs for multi-family malware classification. Due to bias-variance trade off, we select the model trained at the 13-th epoch and apply it on the test data. }
\end{figure}

\begin{figure}
\centering
\includegraphics[width=0.4\textwidth]{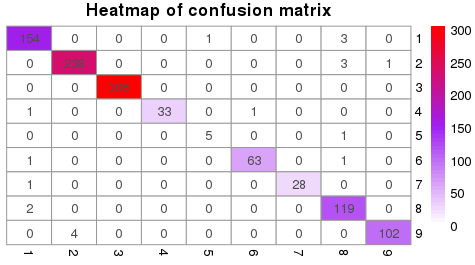}
\caption{\label{fig:confusion_matrix_kaggle}The confusion matrix of our method on test set. The accuracy, average false positive rate and average true positive rate are  98.13\%, 0.237\% , and 96.63\% respectively, outperforming all other selected methods.}
\end{figure}

We train the network for 25 epochs and plot the training and validation accuracy in Figure \ref{fig:training_validation_acc_kaggle}. Due to the bias-variance trade off\cite{duda2012pattern}, we select the model trained at the 13-th epoch and apply it on the test data. On the test set, the classification accuracy by our proposed method is 98.13\% with average false positive rate at $0.237\%$ and average true positive rate at $96.63\%$ respectively. These three metrics outperform all the other 17 classifiers as shown in Table \ref{tab:dataset2}. The confusion matrix is plotted in Figure \ref{fig:confusion_matrix_kaggle}.

The vectorized dimension of images $\in [0,1]^{224\times224}$ is 50176, which is much greater than the number of training samples. Due to the curse of dimensionality \cite{duda2012pattern}, we first apply PCA on the training set and select the first 80 PC dimensions by the scree plot. Then the PCA rotation is applied on the test matrix to obtain the low-rank representation. Again, XGB performs the best second after our method with accuracy of 96.01\%, average FPR at 0.514\% and average TPR at 85.80\%, while LDA performs the worst with accuracy of 76.45\%, average FPR at 3.023\% and average TPR at 63.86\%.

Training from scratch is done using small Inception neural network where the images are resized to 28 by 28 with single channel, since we do not have enough large dataset to enable training from scratch using the original inception network architecture. The resulted accuracy, average FPR and average TPR are 94.57\%, 0.683\% and 89.02\% respectively, which are less effective compared to our method. We also apply the selected classical algorithms on the images vectorized into one-dimensional arrays of length 784. SVM-linear and SVM-radial have similar performance but less effective compared with training from scratch. %Surprisingly, the performance of 5NN degrades significantly. Resizing and $L_2$ distance could be contributing to the degradation. 

%We note the performance of the other classifiers in comparison might improve if we use feature engineering methods rather than image representation. However this again highlights the advantage of deep learning of saving cost at manual feature engineering.
\begin{table*}[t]
\begin{center}
\begin{tabular}{ |l|c|c| c| l| } 
 \hline
Algorithm & Accuracy & $\overline{FPR}$ & $\overline{TPR}$ & Data shape\\ \hline 
Our method & \textbf{98.13\%} & \textbf{0.237\% } & \textbf{96.63\%} &  $224 \times 224 \times 3$ \\ 

TFS via shallow NN $\circ$ PCA & 82.41\% & 2.551\% & 59.38\% & $224 \times 224 \times 1 \rightarrow 50176 \rightarrow 80$\\ 

Naive Bayes$\circ$ PCA  & 74.23\%& 3.116\% &73.06\%&$224 \times 224 \times 1 \rightarrow 50176 \rightarrow 80$\\ 
5-nearest neighbor$\circ$ PCA  & 95.31\%&  0.602\% & 85.75\%&$224 \times 224 \times 1 \rightarrow 50176 \rightarrow 80$\\ 
LDA $\circ$ PCA & 76.45\% & 3.023\%  & 63.86\% &$224 \times 224 \times 1 \rightarrow 50176 \rightarrow 80$ \\ 
Random forest$\circ$ PCA  & 95.73\%& 0.548\% &  84.26\%&$224 \times 224 \times 1 \rightarrow 50176 \rightarrow 80$ \\ 
XGB $\circ$ PCA & 96.01\%& 0.514\%  &  85.80\%&$224 \times 224 \times 1 \rightarrow 50176 \rightarrow 80$\\ 

SVM-linear$\circ$ PCA  & 86.35\% & 1.799\%  & 72.71\% &$224 \times 224 \times 1 \rightarrow 50176 \rightarrow 80$\\ 
SVM-radial$\circ$ PCA  & 86.26\% & 1.975\% &  72.14\%&$224 \times 224 \times 1 \rightarrow 50176 \rightarrow 80$ \\ 

%%%%%%%%%%%%% 28 by 28
TFS via Small Inception & 94.57\% & 0.683\% & 89.02\% &   $28 \times 28 \times 1$\\ 
TFS via shallow NN & 82.32\% & 2.382\% &67.01\% &$28 \times 28 \times 1 \rightarrow 784$\\ 
Naive Bayes & 72.09\%&  3.563\% &62.93\%&$28 \times 28 \times 1 \rightarrow 784$\\ 
5-nearest neighbor & 72.51\%& 3.879\% & 55.89\%&$28 \times 28 \times 1 \rightarrow 784$\\ 
LDA $\circ$ PCA & 72.37\% &3.609\%  & 59.72\% &$28 \times 28 \times 1 \rightarrow 784$\\ 
Random forest & 86.35\%& 1.845\% &  73.05\%&$28 \times 28 \times 1 \rightarrow 784$ \\ 
XGB & 90.15\%&  1.299\%  &  79.00\%&$28 \times 28 \times 1 \rightarrow 784$\\ 

SVM-linear & 92.14\% & 0.379\%  & 78.12\% &$28 \times 28 \times 1 \rightarrow 784$ \\ 
SVM-radial & 92.25\% & 0.374\% &  78.36\% & $28 \times 28 \times 1 \rightarrow 784 $\\ 
\hline
\end{tabular}
\end{center}
\caption{Comparison of algorithm performance on Dataset II. Again, our proposed method achieves the highest classification accuracy, highest average true positive rate and lowest false positive rate. }
\label{tab:dataset2}
\end{table*}

%\begin{figure}
%\centering
%\includegraphics[width=0.5\textwidth]{Figs/train_df_scree_plot.png}
%\caption{\label{fig:confusion_ucsb}The confusion matrix of applying our proposed method on the Kaggle dataset. The majority of the malware samples are correctly classified, achieving accuracy of (fill in). }
%\end{figure}

\subsection{Binary Classification}
\begin{figure}
\centering
\includegraphics[width=2in, height=.5in]{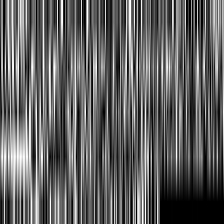}
\caption{\label{fig:bad_file_example}We remove files with size less than 5 kb, since after reshaping, the patterns are greatly distorted.}
\end{figure}

Our third experiment is a binary classification to distinguish benignware and malware. 
After calculating the size distribution, we remove the benign files under 5kb, since the patterns in the image are greatly distorted after resizing as seen in Figure \ref{fig:bad_file_example}. After removal, the dataset consists of 16518 benign files and 10639 malicious files.

 %When applying transfer learning via Inception-v1 and training from scratch via small Inception, the training:validation:test ratio is set at 0.8:0.1:0.1. For the classical machine learning algorithms, the training and testing ratio is set at 0.8:0.2. 

\begin{figure}
\centering
\includegraphics[width=0.5\textwidth]{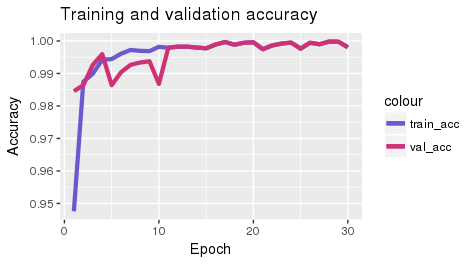}
\caption{\label{fig:binary_training_val_acc}The training and validation accuracy of our method during 25 epochs for binary classification. To avoid overfitting, we use the model at the 14-th epoch for testing.}
\end{figure}

\begin{figure}
\centering
\includegraphics[width=0.27\textwidth]{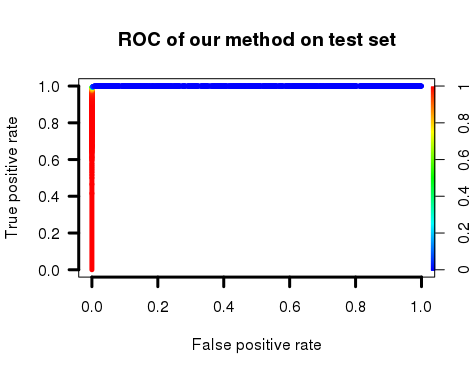}
\includegraphics[width=0.22\textwidth]{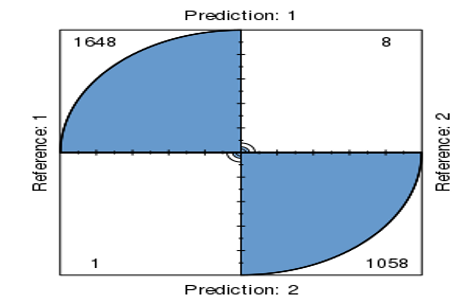}

\caption{\label{fig:roc_binary}Top: The ROC curve of applying our method to the benign-ware and malware test set. The area under the curve is 1. At accuracy of 98.03\%, the false positive rate is 0. Bottom: The confusion matrix of applying our method to the benign-ware and malware test set. The accuracy is 99.67\% with 0.75\% false positive rate, $99.94\%$ true positive rate and 99.73\% $F_1$ score.}
\end{figure}

We fine-tune the network for 25 epochs and plot the training and validation accuracy in Figure \ref{fig:binary_training_val_acc}. We select the model trained at the 14-th epoch and apply it on the test data. On the test set, the classification accuracy by our proposed method is 99.67\% with false positive rate at $0.750\%$, true positive rate at $99.94\%$ and $F_1$ score at 99.73\% respectively. Evaluated in all the four metrics, our method outperforms all the other 17 classifiers as shown in Table \ref{tab:dataset3}. The ROC curve, with area under the curve being 1, and the confusion matrix are plotted in Figure \ref{fig:roc_binary}. At accuracy of 98.03\%, the false positive rate is 0.

%We see that evaluated by all the metrics such as classification accuracy, false positive rate, true positive rate and $F_1$ score, our proposed method achieves the best performance in comparison with the other selected classifiers. The second best performing classifier is training from scratch small Inception network. 

\begin{table*}[t]
\small
\begin{center}
%\resizebox{.7\textwidth}{!}{%

\begin{tabular}{ |l|c|c| c|c|l| } 
 \hline
Algorithm & Accuracy & $FPR$ & $TPR$ & $F_1$ &  Data shape\\ \hline 
Our method & \textbf{99.67\%} & \textbf{0.750\% } & \textbf{99.94\%} & \textbf{99.73\%} & $224 \times 224 \times 3$ \\

TFS via shallow NN $\circ$ PCA & 63.59\% & 92.95\% & 100.0\% & 76.96\%&  $224 \times 224 \times 1 \rightarrow 50176 \rightarrow 80$\\ 

Naive Bayes$\circ$ PCA  & 48.82\%& 6.770\% &20.22\%& 32.47\% & $224 \times 224 \times 1 \rightarrow 50176 \rightarrow 80$\\ 
5-nearest neighbor$\circ$ PCA  & 98.16\%& 2.962\% & 98.88\% & 98.49\%&$224 \times 224 \times 1 \rightarrow 50176 \rightarrow 100$\\ 
LDA $\circ$ PCA & 87.70\% & 23.60\%  & 94.97\% & 90.38\%&$224 \times 224 \times 1 \rightarrow 50176 \rightarrow 100$ \\ 
Random forest$\circ$ PCA  & 98.05\%& 3.150\% &  98.82\%& 98.40\%&$224 \times 224 \times 1 \rightarrow 50176 \rightarrow 100$ \\ 
XGB $\circ$ PCA &98.16\%& 2.774\%  &  98.76\%&98.49\%& $224 \times 224 \times 1 \rightarrow 50176 \rightarrow 100$\\ 

SVM-linear$\circ$ PCA  & 90.83\% & 15.70\%  & 95.03\%& 92.65\% &$224 \times 224 \times 1 \rightarrow 50176 \rightarrow 100$\\ 
SVM-radial$\circ$ PCA  & 97.11 \% & 6.91\% &  99.70  \%&97.67\% &$224 \times 224 \times 1 \rightarrow 50176 \rightarrow 100$ \\ 

%%%%%%%%% 28 by 28 to 784 
TFS via Small Inception & 98.42\% &  2.158\% & 98.79\% &  98.70\% & $28 \times 28 \times 1$\\

TFS via shallow NN & 93.11\% &  7.475\% &93.49\% &94.29\%  &$28 \times 28 \times 1 \rightarrow 784$\\ 
Naive Bayes & 79.83\%& 35.87\% &89.95\%&84.44\%& $28 \times 28 \times 1 \rightarrow 784$\\ 
5-nearest neighbor & 80.52\%& 12.60\% & 76.08\%&82.61\%& $28 \times 28 \times 1 \rightarrow 784$\\ 
LDA $\circ$ PCA & 85.64\% &29.71\%  & 95.52\% &89.00\%& $28 \times 28 \times 1 \rightarrow 784 \rightarrow 50$\\ 
Random forest &  92.78\%& 17.44\% &  99.36\%&94.36\%& $28 \times 28 \times 1 \rightarrow 784$ \\ 
XGB & 94.92\%& 9.450\%  &  97.73\%&95.90\%& $28 \times 28 \times 1 \rightarrow 784$\\ 

SVM-linear & 84.48 \% & 25.06\%  & 90.61\% & 87.66\% &  $28 \times 28 \times 1 \rightarrow 784$\\ 
SVM-radial & 93.26\% & 14.06\% &  97.97\% &94.65\% & $28 \times 28 \times 1 \rightarrow 784$ \\ 
\hline
\end{tabular}
\end{center}
\caption{Comparison of algorithm performance on Dataset III. Our proposed method achieves the highest classification accuracy, highest average true positive rate, lowest false positive rate and highest $F_1$ score. }
\label{tab:dataset3}
\end{table*}

%\begin{figure}
%\centering
%\includegraphics[width=0.4\textwidth]{Figs/binary_confusion_matrix_plot.png}
%\caption{\label{fig:confusion_binary}The confusion matrix of applying our method to the benign-ware and malware test set. The accuracy is 99.67\% with 0.75\% false positive rate, $99.94\%$ true positive rate and 99.73\% $F_1$ score.}
%\end{figure}

In summary, all three experiments demonstrate the classification efficacy of our proposed methodology superior to the selected classical machine learning algorithms and training from scratch.

\subsection{Prediction interpretation}
Simply trusting the best performing classifier is equivalent to putting blind faith in the black-box ML models. Our proposed approach addresses the trustworthy issue of  deep transfer learning for image-based malware classification. Here we illustrate via one prediction how interpretation can be generated and can be helpful for malware practitioners to triage malware or discover new signatures.

%While such method achieves superior performance compared to classical machine learning methods and also requires almost no manual feature engineering step, one aspect – this proposed method’s interpretability and trustworthiness was not explored, defined and discussed. In fact, this is a general issue in machine learning applications and deployment. The ML models are mostly black-box without much explainable or even intelligent interpretability. Simply trusting the best performing classifier is equivalent to putting blind faith in the black-box ML models. 

%a trust component, to enhance the trust of the image-based deep transfer learning algorithm for malware classification. This explanation scheme via solving an optimization problem helps explain the prediction of deep learning based malware classifier. With such information, security practitioners build up confidence on deploying and integrating the interpretable and faith model into their defense product. 

\begin{figure}
\centering
\includegraphics[width=0.5\textwidth]{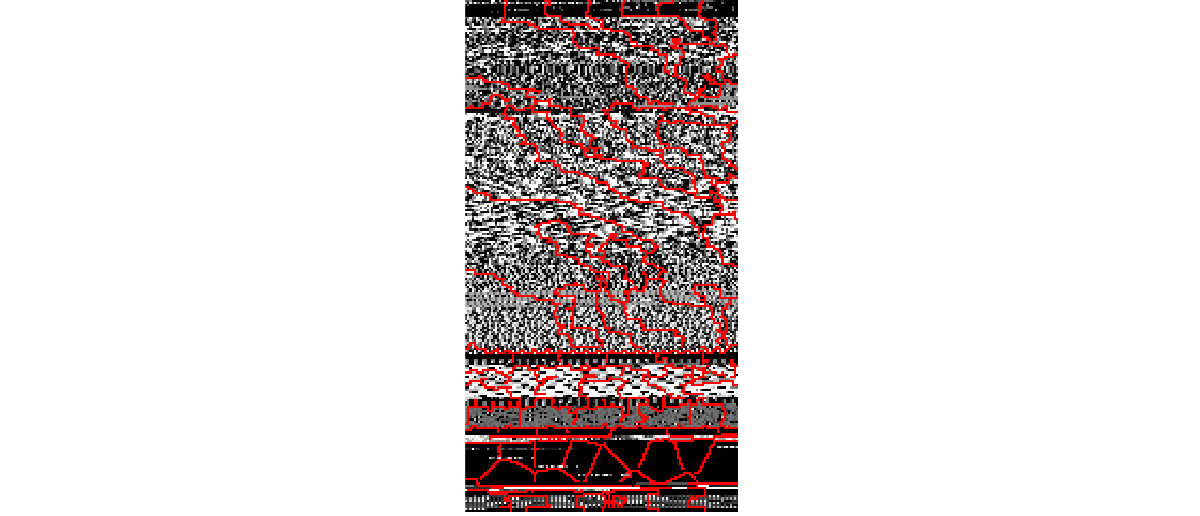}
\caption{\label{fig:Super_pixel} The malware image is represented as 200 super pixels with boundaries drawn in red.}
\end{figure}

\begin{figure}
\centering
\includegraphics[width=0.5\textwidth]{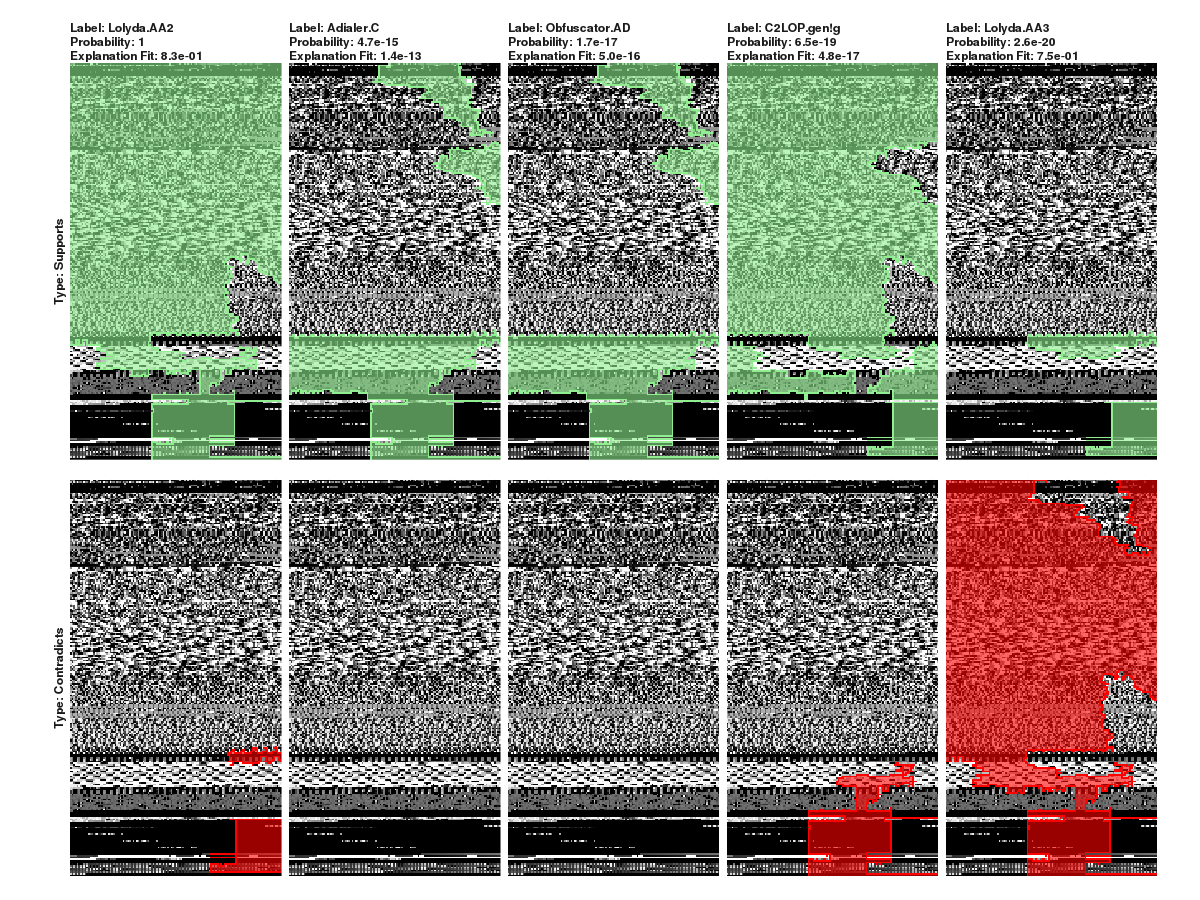}
\caption{\label{fig:interpretation_sample} Visual interpretation of what the algorithm sees. The red regions indicate the pixel regions which the model does not believe they contribute to the prediction. Most area are plotted as green. On the other hand, the top 5-th prediction is Lolyda.AA3, and most of the regions are red, indicating the model sees the least of Lolyda.AA3 family in this malware image.}
\end{figure}

%\begin{figure}
%\centering
%\includegraphics[width=0.5\textwidth]{Figs/process.png}
%\caption{\label{fig:trust_component} The process of the trust component}
%\end{figure}
We select a malware image from malware family Lloyda.AA2 and plot 200 super-pixel representations in Figure \ref{fig:Super_pixel}.
Each super pixel is associated with a feature value of 0 indicating absence and 1 otherwise.
%%%%% rewrite
%Our model predicts the sample with over 99.9\% probability, which means the classification is both correct and confident. 
We examine what aspects of the malware image are considered by the model to produce the prediction and plot the top 5 classes predicted by the model as seen in Figure \ref{fig:interpretation_sample}.
The top 1 predicted class is Lolyda.AA2 with prediction probability above 99.9\%. In the far left plot in Figure \ref{fig:interpretation_sample}, the green regions indicate the pixel regions where the model believes they are contributing to the prediction decision as Lolyda.AA2. The red regions indicate the pixel regions where the model does not believe they contribute to the prediction. Most area are plotted as green. On the other hand, the top 5-th prediction is Lolyda.AA3, where most of the regions are red, indicating the model sees the least of Lolyda.AA3 family in this malware image. Security practitioners can use such visual explanation and interpretation to identify new malware patterns and boost trust in deep learning models for image-based malware classification.

%An interpretable representation is a binary vector of value {0, 1} to indicate the existence of the super-pixel region/patch. Then a binary vector of {0, 1}-values, where 0 indicates the absense of the super-pixel and 1 indicates the presence of the super-pixel, is associated with each of the super-pixel. We use sparse linear classifiers to train on the binary vectors and obtain the trained coefficients of the function. 

%Plotted in red regions are the positive coefficients associated with each super pixel. For the correct Class Lloyd.AA2, most regions of 

%Notice green areas are also a lot in another class. When we do the examination the textural are similar. 

%means the regions of pixels are contributing to the classification decision by the model. The negative coefficients associated with the super pixels mean these regions in the malware image do not contribute to the classification decision.

%\begin{figure}
%\centering
%\includegraphics[width=0.5\textwidth]{Figs/AI_explanation.png}
%\caption{\label{fig:Super_pixel} The superpixel representation of the malware images}
%\end{figure}

\section{Discussion and Conclusion} \label{sec:disc}
In this work we propose using transfer learning from computer vision for static malware classification. Via three datasets, we demonstrated the superior performance of the proposed technique in comparison with training from scratch and other classical machine learning algorithms. We discuss our future research agenda. 

\subsection{Convex Combination for Enhanced Malware Classification}
%While transfer learning is suitable for borrowing domain knowledge from natural images, to extend to our target domain malware analysis, we conjecture some features extracted using transfer learning training scheme might be more biased towards natural images. On the other hand, 
Unlike transfer learning which only fine-tunes a small portion of layers in a deep neural network, training from scratch convey information from the malware images within all the layers. Furthermore as seen in our third experiment, we removed smaller-sized images due to the skewness of the file size distribution. 
The above issues motivate us to consider a convex combination of training from scratch and transfer learning, where transfer learning is utilized on the high-resolution images and training from scratch is applied on the low-resolution images. The formulation of such convex combination is as follows:

\begin{equation*}
\begin{aligned}
& \underset{\alpha, \beta}{\text{maximize}}
& & \mathcal{F}(\alpha M_1 + \beta M_2) \\
& \text{subject to}
& & \alpha + \beta = 1.
\end{aligned}
\end{equation*}
The function $\mathcal{F}$ is a pre-specified evaluation metric such as accuracy, true positive rate, $F_1$ score, area under the curve or the negative of false positive rate.
Our preliminary experiment shows that convex combination scheme outperforms transfer learning or training from scratch alone. It reduces the false positive rate by 4.2 times while increasing the accuracy by 0.3\%. We plan to conduct thorough investigation for the convex combination of transfer learning and learning from scratch for enhanced static malware analysis.

\subsection{Trust Score for Image-based Malware Detector}
We will continue the study of interpretability and explainability of deep learning models for image-based malware detection. %We can utilize the interpretation component to provide visual understanding of the malware images and map it back to specific locations of the binaries to study. 
We will investigate the scheme of establishing an overall trustworthy score for the deep learning model and use such a score for model selection for deployment in cyber-security applications.

\subsection{Resiliency and Adversarial Machine Learning}
While our proposed method is applicable on malware with or without obfuscation, it is worthwhile to study whether the methodology is resilient against code obfuscation. 
%how code obfuscation affects the classification accuracy and whether the methodology can be resilient on such adversarial examples.
%Our system is resilient to obfuscation. When malware from the same family are packed with the same packer, the images of the packed malware appear similar. To avoid packers, some families directly embed a polymorphic engine into the code, for example, section encryption in several layers. Using very deep neural network is able to extract the textural information preserved by such weak encryption schemes.   
Recent studies show that small adversarial pixel perturbations on images, even unnoticeable to the human eyes, can lead to misclassification by deep learning image classification \cite{goodfellow2014explaining}\cite{moosavi2017universal}\cite{papernot2016transferability}. However via adversarial training, the underlying deep learning algorithms become more robust against adversarial samples. We plan to study how to best employ adversarial perturbation on natural images to malware images and utilize adversarial training to create strong defense.
%We will also study adversarial perturbations can be useful in terms of creating robust defense against adversarial attacks. 

\section*{Acknowledgement}
The author would like to thank Ravi Sahita and David Durham for their valuable feedback and discussion.
\bibliographystyle{IEEEtran}
\bibliography{references}
\end{document}